\documentclass{article} 
\usepackage[]{aaai23} 
\nocopyright
\usepackage{times}  
\usepackage{helvet}  
\usepackage{courier}  
\usepackage[hyphens]{url}  
\usepackage{graphicx} 
\usepackage{natbib}  
\usepackage{caption} 
\usepackage{algorithm}
\usepackage{algorithmic}

\usepackage{subcaption}
\usepackage{newfloat}
\usepackage{listings}
\DeclareCaptionStyle{ruled}{labelfont=normalfont,labelsep=colon,strut=off} 
\floatstyle{ruled}
\newfloat{listing}{tb}{lst}{}
\floatname{listing}{Listing}
\usepackage{amsfonts,amsmath,algorithm}
\usepackage{amssymb}
\usepackage{bm}
\usepackage{mathtools}
\usepackage{mathabx}

\usepackage{graphicx}
\usepackage{caption}
\usepackage{float}

\usepackage{hyperref}
\usepackage{xspace}
\usepackage{multirow}
\usepackage{physics}
\usepackage{cancel}
\usepackage{txfonts}

\RequirePackage[textsize=scriptsize]{todonotes}

\usepackage{thmtools}
\usepackage{thm-restate}

\renewcommand{\H}{\mathbf{H}}

\newcommand{\vihari}[1]{} 

\NewDocumentCommand{\tens}{e{_^}}{%
  \mathbin{\mathop{\otimes}\displaylimits
    \IfValueT{#1}{_{#1}}
    \IfValueT{#2}{^{#2}}
  }%
}

\DeclareMathOperator*{\argmin}{argmin}

\newtheorem{rem}{Remark}

\newcommand{\indep}{\bot\!\!\!\!\bot}

\newcommand{\LP}{\mathcal{L}_\text{Prim}}
\newcommand{\LA}{\mathcal{L}_\text{Aux}}

\newcommand{\rom}[1]{\uppercase\expandafter{\romannumeral #1\relax}}

\makeatletter
\newcommand*{\centernot}{%
  \mathpalette\@centernot
}
\def\@centernot#1#2{%
  \mathrel{%
    \rlap{%
      \settowidth\dimen@{$\m@th#1{#2}$}%
      \kern.5\dimen@
      \settowidth\dimen@{$\m@th#1=$}%
      \kern-.5\dimen@
      $\m@th#1\not$%
    }%
    {#2}%
  }%
}
\makeatother

\makeatletter
\DeclareRobustCommand\notindep{\mathrel{\m@th\mathpalette\c@ncel\indep}}
\makeatother

\title{Implicit Training of Energy Model for Structured Prediction}
\author {
     Shiv Shankar,
     Vihari Piratla
 }
 \affiliations {
     sshankar@cs.umass.edu, 
     vp421@cam.ac.uk
 }

\begin{document}

\maketitle

\begin{abstract}


Much research in deep learning is devoted to developing new model and training procedures. On the other hand, training objectives received much less attention and are often restricted to combinations of standard losses. When the objective aligns well with the evaluation metric, this is not a major issue. However when dealing with complex structured outputs, the ideal objective can be hard to optimize and the efficacy of usual objectives as a proxy for the true objective can be questionable. In this work, we argue that the existing inference network based structured prediction methods~\citep{tu-18, tu2020improving} are indirectly learning to optimize a dynamic loss objective parameterized by the energy model. We then explore using implicit-gradient based technique to learn the corresponding dynamic objectives. Our experiments show that implicitly learning a dynamic loss landscape is an effective method for improving model performance in structured prediction.



%
\end{abstract}

\section{Introduction}
Deep neural networks have achieved widespread success in a multitude of applications such as translation \citep{vaswani2017attention}, image recognition \citep{he2016identity}, objection detection \citep{ren2016faster} and many others. This success has been enabled by the development of backpropagation based algorithms, which provide a simple and effective way to optimize a loss calculated on the training set.
Generally a large portion of existing work has focused only on designing of models and optimization algorithms. However with the increased prevalence of meta-learning, researchers are exploring new loss objectives and training algorithms~\citep{wu2018learning,huang2019addressing, oh2020discovering}.

Intuitively one would like to choose objectives which can dynamically refine the kind of signals it produces for a model to follow, in order to guide the model towards a better solution. Oftentimes standard objectives are pretty effective at this; however, these objectives have generally been explored for simple predictions. When dealing with complex outputs, there is a significant scope for improvement by designing better training objectives. A good example is structured prediction \citep{ belanger2016structured}, where the output includes multiple variables and it is important to model their mutual dependence.  One natural candidate is to use the likelihood under a probabilistic model that captures this dependence. Such models though cannot be used to efficiently predict the output and require inference. On the other hand, one can reduce inference complexity by simplifying the modeled dependencies \citep{Lafferty:2001:CRF:645530.655813}. 

An ideal loss function in this case would naturally guide the model towards incorporating the output correlations while allowing a more standard feed-forward or similar predictive model to quickly and efficiently produce the output. Energy based structured prediction \citep{belanger2016structured} provide a natural framework in which one can explore learned losses by using the energy itself as the training objective. Existing works \citep{tu-18,tu2020improving} have looked at learning inference networks to directly predict structured outputs, and not on the energy based objective itself.

\textbf{Contributions} This work explores the thread of learning dynamic objectives for structured prediction. Using the insight of \citet{hazan2010direct}, we connect the existing paradigm of \citet{tu2020improving} to a surrogate loss learning problem. This allows us to identify a key problem with the approach of \citet{tu2020improving}, that it indirectly changes the surrogate objective problem to an adversarial problem. Next we build on this idea, and propose to use implicit gradients \citep{krantz2002implicit} for learning an energy based structured prediction model. We then use ideas from \citet{christianson1992automatic} to compute gradients at scale for the corresponding optimization problem. The experimental results show the effectiveness of our methods against competitive baselines on three tasks and nine datasets with very large label space.

\section{Preliminaries}


\subsection{Learning a Loss Function}
We describe in this section a formulation for learning a dynamic loss function. This loss function, which we call auxiliary loss is used to train a \emph{model}. The model trained on this auxiliary loss is then evaluated on a different loss function, called the primary loss. This second loss function is called primary loss because this is the 'true' loss of concern to the user. For example, in a standard supervised learning setting,  the primary loss could be the performance of the model on a validation set. The goal then is to learn the auxiliary loss in a way that the learnt model's performance as measured by the primary loss is optimized. 
If the model is denoted by $f_\theta$ with parameters $\theta$, auxiliary loss by $\LA$, and primary loss by $\LP$; then this problem can be written as:
\begin{align}
\begin{aligned}
    &\min_{\LA} \LP(\argmin_\theta \LA(\theta)) 
\end{aligned}
\end{align}
Variants of the same formulation have been explored for supervised learning \citep{huang2019addressing,wu2018learning} and reward learning \citep{Bechtle19,zheng2018learning}.
The outer problem is technically a problem of optimization over the space of functions. To be able to solve this computationally, the auxiliary loss $\LA$ is parameterized with some parameters $\phi$. The problem of learning $\LA$ can then be changed to learning $\phi$ as:

\begin{align}
\label{eqn:prob_form}
\begin{aligned}
    &\phi^* = \argmin_\phi \LP(\theta^*(\phi), \phi) \\
    &\text{such that} \\
    &\theta^*(\phi) = \argmin_\theta \LA(\theta, \phi)
\end{aligned}
\end{align}


\subsection{Implicit Gradient Method}
The aforementioned problem is a bi-level optimization problem. In such a case, a parameter ($\phi$) that influences $\LA$, can influence the primary objective $\LP$ via the dependence of the inner optimized parameters $\theta*$ on $\phi$.
The implicit gradient method \citep{krantz2002implicit,dontchev2009implicit} provides a way to compute the gradient of $\LP$ wrt $\phi$ due to this implicit dependence.

For the problem given in Equation \ref{eqn:prob_form}, under certain regularity conditions, $\LP$ is a differentiable function of $\phi$ and its gradient is given by:
\small
\begin{align}
\frac{\partial \LP}{\partial \phi} &= \underbrace{\frac{\partial (\LP(\theta^*(\phi), \phi) )}{\partial \phi}}_{\text{Explicit gradient}} - \\
&\underbrace{\left[ \frac{\partial}{\partial \phi} \frac{\partial}{\partial \theta}(\LA(\theta,\phi)) \right] \left[ \frac{\partial}{\partial \theta} \frac{\partial}{\partial \theta}(\LA(\theta,\phi)) \right]^{-1} \frac{\partial ( \LP(\theta^*(\phi), \phi) }{\partial \theta}}_{\text{Implicit Gradient}} 
\label{eqn:grad_thm}
\end{align}
\normalsize
The existence of gradient follows from Theorem 2G.9 in \citet{dontchev2009implicit}. The derivation of the Equation~\ref{eqn:grad_thm} is presented in the Appendix~\ref{apx:meta}. 
As can be seen from the above equation, the true gradient has two terms: a standard component $\partial_\phi \LP$ and the implicit component due to the the dependence of optimal $\theta$ on $\phi$. We will sometimes abuse terminology to call this term as implicit or meta gradient.

\section{Related Work}
\paragraph{Implicit Gradients} Implicit gradients are a powerful technique with a wide range of applications. Recently they have been used for applications like few-shot learning \citep{rajeswaran2019meta, lee2019meta} and building differentiable optimization layers in neural-networks \citep{amos2017optnet,agrawal2019differentiable}. These techniques also arise naturally in other problems related to  differentiating through optimizers \citep{vlastelica2019differentiation}, such as general hyper-parameter optimization \citep{lorraine2020optimizing}. For more detailed review of implicit gradients we refer the readers to \citet{dontchev2009implicit, krantz2002implicit}. Implicit gradient methods have been used for energy based learning of MRFs \citep{tappen2007learning, samuel2009learning}. These works were further extended to use finite-difference methods. While, our work is similar in that it focuses on using implicit gradients for learning energy based models; we focus on structured prediction tasks.

\paragraph{Structured Prediction}
\label{sec:rel_sp}
While structured prediction has been a well studied task \cite{sarawagi2008accurate, Lafferty:2001:CRF:645530.655813}; in recent years energy based models have become prominent in the field \citep{belanger2016structured, rooshenas19search, tu2019benchmarking}. These models essentially relax the output space to a continuous version on which an energy function is learnt for scoring the outputs. Structured prediction energy networks \citep{End-to-EndSPEN, rooshenas19search} pair up such energy based models with gradient-based inference for prediction. The training methods for these models have generally relied on generalized version of structural SVM learning \citep{ssvm}, with repeated cost augmented inference being done to adapt the energy models landscape. Due to the difficulty of prediction and instability in training such models \citet{tu-18} propose an approach called InfNet which directly performs the inference step instead of using gradient descent or other optimization procedures. Our work directly builds upon recent research on InfNet \citep{tu2020improving}. The most important difference between these works and ours is the bi-level optimization formulation and use of implicit gradients. To the best of our knowledge no work in structured prediction literature uses implicit gradient based methods. Secondly, most works either use cost-augmented inference during training \citep{rooshenas19search, belanger2016structured} or use the inference network and energy network in an adversarial min-max game \citep{End-to-EndSPEN, tu-18}. The former increases inference time significantly while the latter uses incorrect gradients. ALEN \citep{pan2020adversarial} propose augmenting the deep energy model of a SPEN with adversarial loss.

An important difference of our method differs from these methods is that we 'meta-learn' the energy function as a trainable objective and can be applied to adjust training of these models as well. Moreover models like GraphSPEN which incorporate constrained inference are not scalable. Our approach side-steps this issue by using an Inference Network \citep{tu-18} approach. Finally ideas from energy based learning have been used in translation \citep{tu-etal-2020-engine,bhattacharyya2020energy,edunov2017classical} and text generation \citep{deng2020residual}.

\paragraph{Learning Dynamic and Surrogate Losses} 
Due to the nature of many real-life structured losses (like BLEU, F1, IoU), a natural approach is to build a proxy or surrogate loss. In fact the classic cost sensitive hinge/margin loss used in \citet{ssvm}, is a convex surrogate of the true cost \citep{hazan2010direct}. Similarly the value network method of \citet{gygli2017deep} aims to learn a differentiable energy network which directly predicts the score/task-loss of an output. Multitude of works have focused on constructing efficiently optimizable surrogates \citep{hazan2010direct,song2016training,gao2015consistency}. Surrogate loss learning was formulated as a multi-level optimization by \citet{colson2007overview}. Our work uses the insight of \citet{hazan2010direct}, to interpret learning a structured energy model as a surrogate loss learning problem and uses the bi-level optimization framework to solve the corresponding task.

Modern works such as that of \citet{wu2018learning, huang2019addressing,Bechtle19} have attempted to learn dynamic losses for standard classification and regression tasks. Other works such as \citet{sung2017learning,epg18,boyd2022learning} have also proposed learning a reward function for optimization. While the goal of these works and ours is similar in that we try to 'learn' an objective loss for increasing a model performance, there are multiple key differences between them. First, these works do not look at the implicit gradient. Instead they rely on ‘unrolling’ one/few-step gradient updates in the inner optimization and then backpropagate through those updates. This leads to improper characterization of the model/optimizee parameters induced by the learned loss. Secondly, in the supervised learning based applications the model tries to boost a validation set performance, while in our case we are optimizing the prediction on the training examples via the task loss function available in the structured prediction setting. A final difference is that of the input to the meta objective. While these works focus on stage of training such as training step, learning rates etc. we are dealing with samples from the training set. 


\paragraph{Meta Learning}
The key idea in meta-learning is to make the model 'aware' of the 'learning process' ~\citep{Schmidhuber:87long, ThrunP98}. Meta-learning is commonly used for learning model parameters $\theta$ that can be easily adapted to new tasks~\cite{mendonca2019guided, gupta2018meta}, multi-task transfer learning \citep{metz2018learning}, learning hyper-parameters \citep{franceschi2018bilevel} or learning policies for parameter update \citep{maclaurin2015gradient, l2l, li2016learning, franceschi2017forward, meier2018online, daniel2016learning}. \citet{rajeswaran2019meta} use implicit-gradient based methods in a MAML setting.
The key difference from standard meta-learning is that meta-learning is focused on learning model parameters. This is in contrast to this method which aims to learn a loss function which is distinct from the model.



\section{Structured Prediction with Dynamic Loss}
In this section we provide a brief overview of structured prediction, before we present a bi-level optimization based method for structured prediction. 
structured prediction deals primarily with predicting a multivariate structured output $y$. Multi-label classification, semantic labeling, dependency parsing are some examples of multivariate structured output.
An abstract structured prediction task can be defined as learning a mapping from an input space $\mathcal{X}$ to a exponentially large label space: $\mathcal{Y}$. The quality of a predicted output is determined by the score function $ s: \mathcal{Y} \times \mathcal{Y} \rightarrow \mathbb{R}$. The score function is used to compare the gold output $y \in \mathcal{Y}$ with another output $ y' \in \mathcal{Y} $ and can be interpreted as a measure of how good $ y' $ is compared to $y$. 
Some common scoring functions are BLEU used for translation~\citep{papineni2002bleu}, Hamming Distance for comparing strings, and F1-score for multilabel classification tasks~\citep{kong2011multi}. 



\subsection{Energy based Structured Prediction} 
Under this approach one uses a network $E_{\phi}: \mathcal{X} \times \mathcal{\bar{Y}} \rightarrow \mathbb{R}$ which provides the energy for pairs of inputs $x$ and the outputs $y$.  Here $\mathcal{\bar{Y}} $ refers to a suitable relaxation of $\mathcal{Y}\in \{0, 1\}^L$ to a continuous space: $\mathcal{\bar{Y}} \in [0, 1]^L$. The energy network is  trained to assign the correct output $y$  a lower energy than incorrect outputs. At test time, predictions are recovered for an input $x$ by finding the structure $y$ with the lowest energy \citep{belanger2016structured}. 

Training such a energy network, however, requires inference during training to find the current highly rated negative output $\bar{y}$. To make inference efficient \citet{tu-18, tu2020improving} propose using {\it inference networks} $F_\theta$ and $A_\theta$ to directly predict the output. $F_\theta$ is the cost-augmented inference network that is used only during training. 
The goal of $F$ is to output candidates $\bar{y}$ with low energy that also have a high task loss. These are then used as effective negative samples to update the energy $E_\phi$.  On the other hand the goal of $A_\theta$ is to predict the minimizer of the energy during testing. These networks are trained via the following min-max game: 

\begin{align}
\begin{aligned}
\min_\phi \max_\theta &\underbrace{[s(F_\theta(x), y) - E_\phi(x,F_\theta(x)) +  E_\phi(x,y)]_+}_{A} \\
&+ \lambda \underbrace{[  - E_\phi(x,A_\theta(x)) +  E_\phi(x,y)]_+}_{B}    
\end{aligned}
\label{eqn:prob_orig}
\end{align}

This objective is the sum of the margin-rescaling objective (Term $A$) and ranking objective ($B$) for the two different inference problems. Term $A$ contains the train-time inference problem is for the cost-augmented inference net $F$; while Term $B$ is for the test-time inference problem network $A$.

\begin{rem}
While one can use different parameters $\theta,\theta'$ to parameterize the $F,A$ networks respectively, for a well trained energy model these networks are not very dissimilar in behaviour. Furthermore, \citet{tu2020improving} also found sharing parameters between F and A helpful. Hence to avoid confusion as well as for notational simplicity we jointly represent these as $\theta$.
\end{rem}

\begin{rem} While ideally the inference networks $F,A$ would be predicting binary vectors, in practice they are used to predict soft examples. The energy function can then be updated both on the real valued vectors or by obtaining hard outputs via sampling, rounding or other methods.
\end{rem}

The adversarial nature of the above objective makes learning difficult in practice \citep{salimans2016improved}. \citep{tu2019benchmarking} experimented with various regularizing terms to stabilize training. \citet{tu2020improving} found that 
training can be improved by removing the hinge loss for optimizing $\theta$. This helps stabilize training by removing the purely adversarial nature of the original formulation of \citep{tu-18}. However, using a different objective for gradients of $\phi$ does changes the original min-max problem in Equation ~\ref{eqn:prob_orig}
to an entirely different one.

\subsection{Bi-level Interpretation}

\citet{hazan2010direct} prove that for linear energy models, the term A in Equation \ref{eqn:prob_orig}
is a convex relaxation of the task loss. This suggests that learning energy function through such an optimization is an indirect way to learn a surrogate loss function. Surrogate loss learning can be formulated as a bi-level optimization with the outer optimization over loss function parameters $\phi$ constantly updating itself to provide better feedback to the prediction model (as discussed in Section 2). Under this view the margin loss based training can be interpreted as the following bi-level objective:

$$
\min_{\phi} \LP(\theta(\phi),\phi) \quad \text { s.t. } \quad \theta(\phi) = \argmin_\theta \LA(\phi, \theta) 
$$
where 
$$ \LA(\theta, \phi) = - ( s(F_\theta(x), y) + E_\phi(x, F_\theta(x)) + \lambda  E_\phi(x,  A_\theta(x)) ) $$
and 
\begin{align*}
\LP(\theta, \phi) = [ & s(F_\theta(x), y) - E_\phi(x, F_\theta(x)) + E_\phi(x, y) ]_+ \\ 
&+ \lambda [ -E_\phi(x, A_\theta(x) +  E_\phi(x, y) ]_+
\end{align*}



The interpretation of the training procedure is that at each step, the inference network is trained to predict an incorrect $\bar{y}$ with low energy and then the energy network is updated to guide the inference network to a newer solution. Next we note that under a well trained $E$ one does not need two different networks $A,F$, and so we combine the two of them in the same network. To train this model, we use gradient descent based optimization, however, instead of backpropagating through the gradient steps, we use the implicit gradient method mentioned earlier to obtain the gradients of $\phi$. 

Under our interpretation, the procedure of \citet{tu2020improving} uses biased gradients during update of $\phi$. Specifically since they only use $\partial_\phi \LP(\theta, \phi)$, their gradient for $\phi$ misses the first term in Equation \ref{eqn:grad_thm}, which captures the influence due to the implicit dependence of $\theta$ on $\phi$. 
Moreover the presence of the inverse Hessian in the missing term also provides insight into why the bi-level approach can be superior. Specifically, the condition number of the Hessian is a useful measure of the hardness of an optimization and an ill-conditioned Hessian would cause the missing term to explode, something which the adversarial training process of \citet{tu2020improving} ignores. We present a more detailed discussion of this in the Appendix.


One can observe that in the above optimization for $\theta$, the observed outputs $y$ only appear directly in the score function $s$. If $s$ is not differentiable (which is usually the case), updates to $A_\theta$ relies on the function $E$. However, during initial steps of training, $E_\phi$ would not yet have learned to score the true output correctly. Thus the model $A_\theta$ will receive poor supervision. To alleviate this issue, we add direct supervision from output $y$ in $\LA$. In this case we use the output of $A_\theta$ to construct a distribution which is updated via the cross entropy (CE)/ log-likelihood (MLE) loss.


Putting these changes (i.e. merging of inference networks and addition of MLE loss) together we get the following auxiliary objective:

\begin{align}
\begin{aligned}
\LA = \mathcal{L}_{MLE}(y, A_\theta(x)) + \lambda E_\phi(x, A_\theta(x)) 
\end{aligned}
\label{eqn:new_la}
\end{align}

\noindent
where $\mathcal{L}_{MLE}$ is a log-likelihood/cross-entropy based loss and  $\lambda$ is a hyperparameter.\footnote{If we replace F by A in $\LA$ objective above both energy based terms become same; next since $s$ is not-directly optimizable we replace it for supervision with $\mathcal{L}_{MLE}$}

\begin{rem}
Unlike standard meta-learning problems, where the outer parameter $\phi$ is used as the initialization point of the model, here we can directly use the learned inference network $A_\theta$ for prediction. However, one can refine the final $\theta$ on a validation set, or attempt to refine the output of network $A_\theta(x)$ via gradient descent on the energy $E_\phi$. We do not use the validation set for further refinement in our experiments. 
\end{rem}

\subsection{Scalable computation of the implicit-gradient}
An astute reader might note that computing the gradient given in Equation \ref{eqn:grad_thm} directly requires the Hessians $\frac{\partial}{\partial \theta} \frac{\partial}{\partial \theta}(\LA(\theta,\phi))$ and $\frac{\partial}{\partial \phi} \frac{\partial}{\partial \theta}(\LA(\theta,\phi))$. While computing the Hessians can be compute-intensive if the dimensionality of the parameters $\theta , \phi$ is large; computing the inverse Hessian is prohibitively more so.
An alternative method is to differentiate through the optimization procedure, however that severely limits the number of optimization steps one can conduct. Moreover truncated optimization will induce its own biases \citep{vollmer2016exploration}. 

Fortunately, we do not need to compute any of the two matrices. Instead we only need the vector product of these hessian matrices (HVP) with the gradient $\frac{\partial ( \LP(\theta^*(\phi), \phi) }{\partial \theta} $. Efficiently doing such operations is a well researched area with numerous methods \citep{christianson1992automatic, vazquez2011new, song2022modeling}. 
The given expression can be transformed into first computing a HVP with the cross-Hessian $\frac{\partial}{\partial \phi} \frac{\partial}{\partial \theta}(\LA(\theta,\phi))$, and then into an inverse-Hessian vector product (iHVP) with the Hessian $\frac{\partial}{\partial \theta} \frac{\partial}{\partial \theta}(\LA(\theta,\phi))$.  
For the inverse Hessian, we use the von-Neumann expansion method suggested in \citet{lorraine2020optimizing}. This allows one to convert iHVP with a matrix $H$ to product to a polynomial in HVP using the same matrix $H$ (details in the Appendix). 
Once every requisite operation has been turned to HVP,  we can use auto-differentiation on perturbed parameters (i.e. finite step divided difference approximation).

\subsection{Primary Loss Design}
\label{sec:struct}
An advantage of breaking this problem as a bi-level optimization is that unlike \citep{tu2020improving} where the objectives being used for training $\phi,\theta$ are by construction adversarial, we can now use different objectives for our primary and auxiliary losses. We implicitly already used this fact when we added the binary cross entropy loss to $\LA$, and wrote slightly different form for $\LA$ in Equation \ref{eqn:new_la}. However we also have the freedom to choose the primary loss $\LP$ which can result in different behaviour for the models. In fact structured prediction literature has explored variety of losses for training energy models. We mention a few of these which we work use as $\LP$ in our experiments. In this section we shall often use $\bar{y}$ to denote an element from $\mathcal{Y}$ which is distinct from the true output $y$.

\textbf{Hinge/SSVM Loss}. Early structured prediction models were often trained with a version of the hinge loss adjusted for the score function \citep{ssvm}. In current parlance it is also known as margin loss. This is one of the components of the loss used in \citep{tu2020improving}. It is given by the following equation:
$$\mathcal{L}_\text{SSVM} =\left[ s(\bar{y}, y) - E_\phi(x,\bar{y}) + E_\phi(x, y) \right]_+$$

\textbf{Contrastive Divergence}. Literature in probabilistic inference have proposed various losses to do maximum likelihood estimate of energy models \citep{gutmann2010noise, vincent2011connection}. A common loss for such training is the contrastive-divergence \citep{hyvarinen2005estimation} based loss which uses samples to approximate the log-likelihood of the model. We use a similar loss augmented with the score function $s$ as shown below. 

$$\mathcal{L}_{CD} =  \log \dfrac{\exp(-E_\phi(x,y))}{\sum\limits_{k=0}^K \exp(-E_\phi(x,\bar{y}_k) + s(\bar{y}_k, y))  )}$$
 
where $\bar{y}_{1..K}$ refers to $K$ possible negative (non-true output) samples and $\bar{y}_0 = y$.

During training $\bar{y}$ in the aforementioned objectives gets replaced by the prediction of the inference net $A_\theta(x)$. When multiple values are required (such as for $\mathcal{L}_{CD}$) we obtain them samples by interpreting the continuous output of $A_\theta(x)_j$ as a Bernoulli random variables, and drawing samples from the corresponding distribution.

\begin{rem}
Learnt loss functions have been used in literature for the outer objective \citep{Bechtle19}. However, these are also loss objectives used to train the prediction model ($\LA$ in our notation). In this work, predictions are obtained from the inference network $A_\theta$, which is trained by optimizing the energy function $E$. Hence we call $E$ dynamic loss in the latter sense. 
\end{rem}

Now we are in a position to state our exact proposal to train structured prediction models. Our proposed method is summarized in Algorithm \ref{alg:struct_train}.

\begin{algorithm}
\caption{Implicit Gradient for structured prediction}\label{alg:struct_train}
\begin{algorithmic}[h]
\REQUIRE Energy Network $E_\phi$, Inference Network $A_\theta$,\\
Regularization $\lambda$,  Training Data $\mathcal{D} = {x_i,y_i}$, \\ Inner/Outer Iterations $T_\text{inner},T_\text{outer}$\\
\STATE Sample $\theta_0,\phi_0$ randomly
\FOR{$T_\text{outer}$ iterations}
   \STATE $t = 0$
   \STATE Obtain sample $x,y$ from $\mathcal{D}$
   \STATE $\theta_p \leftarrow \theta_t$
   \FOR{$T_\text{inner}$ iterations }
   \STATE Compute $\LA (\theta_p, \phi_t)$
   \STATE $\theta_p \leftarrow \theta_p - \eta \nabla \LA (\theta_p, \phi_t)$
   \ENDFOR 
   \STATE $\theta_t \leftarrow \theta_p$
   \STATE Compute $\LP (\theta_t, \phi_t)$
   \STATE Compute $ g = \frac{d}{d \phi} \LP (\theta_t, \phi_t)$ via Equation \ref{eqn:grad_thm}
   \STATE $\phi_{t+1} \leftarrow \phi_{t} - \eta g$
   \STATE $\theta_{t+1} \leftarrow \theta_{t}$
   \STATE $ t \leftarrow t+1$
\ENDFOR
   
\STATE Return resulting model $A_\theta$
\end{algorithmic}
\end{algorithm}

\section{Experiments}

We evaluate our method using classic structured prediction tasks of multi-label classification (MLC) and sequence-labelling ( POS tagging and NER). 

\textbf{Multi-Label Classification}
We use the following multi-label classification datasets for testing our model: bibtex \citep{katakis2008multilabel}, delicious \citep{tsoumakas2008effective}, eurlexev \citep{mencia2008efficient}.
The performance metric is F1 score , which is also the score function used for training our models. The max-likelihood loss $\mathcal{L}_{\text{MLE}}$ in this case is given by the multi-label binary cross entropy (MBCE). We use the output of $A$ as a vector of Bernoulli variables, and MBCE is then just the sum of logistic losses over the individual components of $y$.
\begin{align*}
\mathcal{L}_{\text{MLE}} = MBCE = \sum\limits_{j=1}^{L} &-y^j \log((A_\theta(x))^j) - (1-y^j)\log(1 - (A_\theta(x))^j)
\end{align*}


For fair comparison with earlier works on these datasets, we used the energy network design of \citet{belanger2016structured}.
The corresponding energy function $E_\phi$ is parameterized as:
$$
E_{\theta} = y^T Wb(x) + v^T\sigma(My)
$$
where the parameters $ \theta$ comprise of  $\{ W, v, M, b \}$. Network $b$ is defined by a multilayer perceptron. A similar multilayer perceptron from the basis of the inference network $A_\theta$.

\begin{table}[]
\centering
\small
\begin{tabular}{|l|l||cccc|}
\hline
\multicolumn{2}{|c}{Method} & \multicolumn{4}{c|}{Dataset}             \\
\hline
     &  & BibTex & Delicious & Eurlexev & Bookmark\\ 
\hline
\hline
\multirow{5}{*}{Slow} & SPEN   & 43.12  & 26.56     & 41.75  & 34.4 \\ 
 & NCE   & 20.12  & 16.97     & 19.50   & - \\ 
 & DVN    & 42.73  & 29.71     & 31.90  & 37.1 \\ 
 & ALEN &  46.4   & - & - &  38.3 \\
 & GSPEN & \textbf{48.6} & - & - & \textbf{40.7} \\
\hline
\multirow{5}{*}{Fast} & MBCE     & 42.47  & 30.12     &
43.25 & 33.8 \\ 
& iALEN &  42.8   & - & - &  37.2 \\
& $\mathcal{L}_\text{SSVM}$  & 44.55  & 30.34     & 42.50  & 37.9\\ %
& $\mathcal{L}_\text{DVN}$    & 44.94  & 28.87     & 42.35 & 38.1   \\ %
& $\mathcal{L}_\text{CD}$     & \textbf{46.21}$\dagger$  & \textbf{35.12}     & \textbf{43.49}  & \textbf{38.5}$\dagger$ \\ 
\hline
\end{tabular}
\caption{Performance of our approach with different objectives (SSVM,CD) compared to standard multi-label classification (MBCE) and energy based models (SPEN,DVN,NCE). Our implicit gradient trained model significantly outperforms the other approaches. * denotes we report results from literature and not our own replication \label{tab:struct_expt_small}. $\dagger$ denotes statistically significant }
\end{table}

We experiment with SPEN \citep{End-to-EndSPEN}, DVN \citep{gygli2017deep}, and an energy model trained by NCE loss \citep{gutmann2010noise,ma2018noise}. As a baseline we also present the results of an MLP trained by standard multi-label binary cross entropy, and ALEN, iALEN \citep{pan2020adversarial} and GSPEN. For our proposed implicit training method, we experiment with different objectives for inner-optimization $\LP$ as described in the section: ``Primary Loss Design''. Our results are presented in Table \ref{tab:struct_expt_small}. 

From the experiments, it is clear that our implicit training approach is superior to most current approaches of using energy based models for structured prediction. Our implicit gradient method gives a boost of upto \emph{5 F1 points} depending on the primary loss objective and the dataset. Furthermore, we also note that ($\mathcal{L}_{SSVM}$, SPEN) and ($\mathcal{L}_{DVN}$, DVN) use the same loss and energy function, and the difference in results is attributable to our proposed implicit training of the inference network.
Next, we also note that the only model that outperforms our proposed method is GraphSPEN/GSPEN, which lacks scalability. For example the running-time of GSPEN on Bib (which is our smallest dataset) is more than 6 times our approach. This is due to the need of computationally hard constrained inference in GSPEN and makes it infeasible on the larger datasets that we experiment with in the next section. Finally we see that the contrastive divergence based objective outperforms the other methods, and so we focus on this objective in our other experiments.
\vihari{Why does GraphSPEN perform comparably or better than ours?}

\textbf{Large Scale Multi-label Modelling}.  To demonstrate that our approach is more scalable and general, we apply our approach on two large text based datasets RCV1 \citep{lewis2004rcv1} and AAPD \citep{yang2018sgm}. Most existing works dealing with structured prediction limit themselves to smaller datasets. The state of the art models on these datasets instead rely on standard max likelihood training. The dependence between labels is usually modeled by novel architectures \citep{zeng2021modeling}, transforming the problem into sequence prediction (SGM) \citep{yang2018sgm} or by adding regularization terms to improve representation (LACO) \citep{zhang2021enhancing}. There are no available energy based baselines on these tasks, partly because of intractability of inference required for energy based structured prediction. We use baselines from the aforementioned works, and use a similar architecture to the smaller MLC task for our energy model, except that our feature networks use pretrained BERT models. We also compare to a CRF based SeqTag baseline that uses BERT to learn label embeddings \citep{zhang2021enhancing}, the seq2seq approach of \citet{nam2017maximizing} and \citet{tsai2020order} which is a RNN based auto-regressive decoder. Our results are presented in Table \ref{tab:struct_expt_large}. It is clear that using energy based method significantly outperforms BERT based models and edges out ahead of other methods which explicitly focus on modeling label dependence. 

\begin{table}[]
\centering
\begin{tabular}{|l||cc|cc|}
\hline
Method  & \multicolumn{2}{c|}{RCV} & \multicolumn{2}{c|}{AAPD} \\ 
\hline
  & Mi-F1 & Ma-F1 & Mi-F1 & Ma-F1 \\ 
\hline
\hline
SGM    &  86.9 & - & 70.2 & - \\
BERT-CE  & 87.1 & 66.7 & 74.1 & 57.2 \\
OCD & - & - &  72.1 & 58.5 \\
Seq2Seq & 87.9 & 66.0 & 69.0 & 54.1 \\
SeqTag & 87.7 & 68.7 & 73.1 & 58.5 \\
LACO   & 88.2 & \textbf{69.1} & 74.7  & 59.1 \\ 
\hline
$\mathcal{L}_\text{CD}$     & \textbf{88.5}$\dagger$  & 68.9 & \textbf{75.6}$\dagger$ & \textbf{59.8}$\dagger$ \\
\hline
\end{tabular}
\caption{Performance of our model on large scale multi-label classification against existing models (SGM, OCD, Seq2Seq, BERT-CE, LACO). Our implicit gradient trained model significantly outperforms or matches other approaches. $\dagger$ denotes statistically significanct scores \label{tab:struct_expt_large}}
\end{table}

\textbf{Named Entity Recognition}. For our experiments we work with the commonly used CoNLL 2003 English dataset \citep{conll20003}. Similar to previous work \citep{Ratinov:2009}, we consider 17 NER labels, and evaluate the results based on the F1 score. 
Following \citet{tu-18}, we design the energy network $E_\phi$ and the inference network $A_\theta$ based on Glove based word embeddings \citep{pennington2014glove}. The text embeddings are then provided to bi-LSTMs to form the features $b(x)$ for the energy function. If we denote by $b(x,t)$ the bi-LSTM output at step $t$, then the energy is :
\begin{align}
\label{eqn:energy-sequence-labeling}
E_{\theta}(x, y) &= \sum_{t=1}^T  \sum_{j=1}^L y_{t,j}U_j^\top b(x,t)  + \sum_{t=1}^T y_{t-1}^\top W y_{t}
\end{align}

The parameters $\theta$ compose of the matrix $W$ and the per label parameter $U_j$, along with the LSTM parameters. Similarly $A_\theta(x)$ can be written as a linear MLP over $b(x)$.

We run our models with two different input feature sets. For the NER version, the input consists of only words and their Glove embeddings. NER+ configuration also provides POS tags and chunk information. 
As baselines we use SPEN \citep{End-to-EndSPEN}, InfNet\citep{tu-18}, InfNet+\citep{tu2020improving} and a cross entropy trained BILSTM baseline. Our results in Table \ref{tab:struct_expt_ner} show that implicit models outperform other existing models. Note in particular that our model with SSVM loss is very similar to the InfNet+\citep{tu2020improving} (with the same losses etc.). The difference between these is a) the final layer in the inference networks $F,A$ are not shared in Infnet+ but are in ours and b) the training procedure is different due to using implicit gradients. If both these models are trained correctly then their final performance should be consistent which seems to be case. Finally similar to the previous experiments, we see improved performance with CD losses.

\begin{table}
\centering
\begin{tabular}{|l||c|c|}
\hline
Models  & NER  & NER+  \\
\hline\hline
BILSTM  & 84.9 & 89.1  \\
SPEN    & 85.1 & 88.6  \\
InfNet  & 85.2 & 89.3  \\
InfNet+ & 85.3 & 89.7  \\
\hline
$\mathcal{L}_\text{CD}$     & \textbf{85.7} & \textbf{90.4}$\dagger$  \\
\hline
\end{tabular}
\caption{Test results for NER and NER+  for different energy based models. SSVM and CD refers to our implicit gradient models. $\dagger$ indicates statistical significance 
\label{tab:struct_expt_ner}}
\end{table}

\textbf{Citation Field Extraction}.  
We conduct experiments on a citation-field extraction task. This is an information extraction task where the goal is to segment a citation text into its constituents as Author, Title, Journal, Date etc. We use the extended Cora citation dataset \citet{seymore1999learning} used in \citet{rooshenas19search}. The citation texts have a max length of 118 tokens, which can be labeled with one of 13 tags. 

We explore this task in the indirectly semi-supervised structured prediction of \citet{rooshenas19search}. In this setting we have a few labeled points, and are also given rules based rewards for the unlabeled samples. However the citation reward loss is based on domain knowledge and is noisy. For the task loss we use token-level accuracy to supplement the reward function. Similarly, the model performance is measured on token-level accuracy. We run this task with 1000 unlabeled and 5, 10, and 50 labeled data points. We compare against GE \citep{mann2010generalized},  RSPEN and SGSPEN \citep{rooshenas19search}, DVN \citep{gygli2017deep} and our contrastive divergence method. Our results are presented in Table~\ref{apx:tab:struct_cite}. 

\begin{table}[!ht]
\centering
\begin{tabular}{|l||c|c|c|c|c|}
\hline
   & GE   & RSPEN & DVN  & SGSPEN & Ours($\mathcal{L}_{CD}$) \\
\hline
\hline
5  & 54.7 & 55.0 & 57.4 & 53.0 & \textbf{55.9}$\dagger$ \\
10 & 57.9 & 65.0 & 60.9 & 62.4 & \textbf{67.8}$\dagger$ \\
50 & 68.0 & 81.5 & 79.4 & 82.6 & \textbf{82.9} \\   
\hline
\end{tabular}
\caption{Comparing performance of our approach on the semi-supervised setting for the citation-field extraction task. Our implicit gradient trained model significantly outperforms the other approaches. $\dagger$ denotes statistical significance \label{apx:tab:struct_cite}}
\end{table}

\paragraph{Time Comparisons}
In Table \ref{tab:time}, we provide the training time and inference time comparison of our method against other methods like SPEN and DVN on multi-label classification datasets. As can be seen the inference time of our proposed method is much better than gradient descent based methods of SPEN and DVN. Moreover , the SOTA GSPEN method takes more than 13s ( $>$ 6 times our approach) for one pass over the bib dataset, highlighting its inefficiency which makes using it infeasible on larger tasks.

\begin{table}[ht!]
\centering
\begin{tabular}{|l||c|c||c|c|}
\hline
          & \multicolumn{2}{|c||}{Training Time} & \multicolumn{2}{c|}{Inference Time} \\
\hline
           & Bib   &  Eurlexev   & Bib   &  Eurlexev   \\
\hline
\hline
SPEN       & 28.2  & 134.5 & 3.8       & 24.5     \\
DVN        & 32.1  & 128.7  & 3.8      & 24.6    \\
\hline
$\mathcal{L}_{CD}$   & 27.7 & 45.6  & 1.8    & 12.1    \\
\hline
\end{tabular}
\caption{Training and inference time (sec/epoch) comparison of our approach against SPEN and DVN. Since the number of parameter update steps for our approach is different per epoch than other models, we have normalized training time/epoch by the number of parameter updates. \label{tab:time}}
\end{table}

\section{Conclusion}
\textbf{Summary}
The primary goal of our work is to learn dynamic losses for model optimization using implicit gradients, in a setting with complex outputs such as in structured prediction. This work uses a bi-level optimization framework for structured prediction that uses a dynamic loss. Then we use implicit gradients to optimize an energy-based model in our proposed framework. We also explore possible designs of these dynamic objectives. Our experiments show our approach outperforms or achieves similar results to existing approaches. Our method tends to be more stable than existing approaches based on inference networks and gradient-based inference. 

\textbf{Limitations and Social Impact} Our contributions are mostly restricted to inference network based structured prediction; and our experiments are mostly textual datasets. structured prediction has also been explored in domains like segmentation and generative modelling, but our experiments are of little insight into those areas. Moreover for applications like translation etc. energy based models have had limited successes, and our approach does not trivially apply to such tasks. Finally even though our approach trains better than other energy based methods, they are still more sensitive to hyperparameters than standard autoregressive or CRF based models. We do not foresee any negative societal impact from this work.









\bibliography{mybib}

\onecolumn
\setcounter{thm}{0}

\appendix

\section {Implicit Gradient}
\label{apx:meta}

Consider the following bi-level objective
\begin{align}
    &\phi^* = \argmin_\phi \LP(\theta*(\phi), \phi) \text{    s.t.   }
    \theta^*(\phi) = \argmin_\theta \LA(\theta, \phi)
\end{align}

Here we have explicitly added the dependence of $\theta$ due to the opimization process on $\phi$. 
One approach to find the optimal $\phi^*$ is to find the partial derivative of $\mathcal{L}(f(\theta^*(\phi)), D_{\text{val}})$ with respect to the $\phi$, and use gradient descent based optimization. 
The corresponding partial derivative is given by 
\begin{align}
 \frac{d}{d \phi} \LP(\theta^*(\phi), \phi) =  \underbrace{   \frac{\partial }{\partial \theta} ( \LP(\theta^*(\phi), \phi) )|_{\theta^*}   }_{1}    \circ \underbrace{  \frac{\partial}{\partial \phi} \theta^*   }_{\rom{2}} + \frac{\partial}{\partial \phi} \LP(\theta^*(\phi), \phi)
 \label{eqn:11}
\end{align}

Considering that the inner optimization is finished we have direct access to $\theta^*(\phi)$ and the first term $\rom{1}$ in the previous equation can be computed directly. The second term is a more challenging to compute.

Implicit gradient method computes this gradient via differentiation of the optimality criteria of the inner optimization. The optimality criteria states that the gradient of the inner loss at the optima $\theta^*(\phi)$ is zero i.e.

\begin{align}
&\nabla_\theta \LA(\theta, \phi) = 0 \Rightarrow \frac{\partial}{\partial \theta} (\LA(\theta, \phi)) = 0
\end{align}

By differentiating this with respect to $\phi$ one gets:
\begin{align}
&\frac{d}{d \phi} \frac{\partial}{\partial \theta} \LA(\theta,\phi) = 0 \Rightarrow \frac{\partial}{\partial \phi} \frac{\partial}{\partial \theta}(\LA(\theta,\phi)) + \frac{\partial}{\partial \theta} \frac{\partial}{\partial \theta}(\LA(\theta,\phi))\frac{\partial}{\partial \phi} \theta   = 0\\
&\rightarrow \frac{\partial \theta}{\partial \phi} = - \left[ \frac{\partial}{\partial \phi} \frac{\partial}{\partial \theta}(\LA(\theta,\phi)) \right] \left[ \frac{\partial}{\partial \theta} \frac{\partial}{\partial \theta}(\LA(\theta,\phi)) \right]^{-1} 
\end{align}

Putting this back in Equation \ref{eqn:11} we get

\begin{align}
\label{eqn:implireg1}
 \frac{d}{d \phi} \LP(\theta^*(\phi), \phi) =  - \left[ \frac{\partial}{\partial \phi} \frac{\partial}{\partial \theta}(\LA(\theta,\phi)) \right] \left[ \frac{\partial}{\partial \theta} \frac{\partial}{\partial \theta}(\LA(\theta,\phi)) \right]^{-1} \frac{\partial ( \LP(\theta^*(\phi), \phi) }{\partial \theta}
     + 
 \frac{\partial (\LP(\theta^*(\phi), \phi) )}{\partial \phi} 
\end{align}

\begin{rem}
Notice that in Equation \ref{eqn:implireg1}, the inverse of second order derivative i.e. a Hessian  needs to be computed which can be expensive. In practice, approximations via Conjugate Gradient, Shermon-Morrison Identity, diagonalized Hessian or von-Neumann Expansion could be used. For our structure prediction experiments we used the von-Neumann approximation.
\end{rem}

\subsection{Approximation for inverse Hessian}
\label{apx:approx}
\paragraph{von Neumann Approximation}
When we apply the von Neumann series for inverse operators on the matrix $I-H$ we get:
$$\\H^{-1} = (I - (I - H))^{-1} = \sum \limits_{i=0}^{\infty} (I - H)^i \approx \sum \limits_{i=0}^{K} (I - H)^i $$
This is convergent for matrices $H$ with singular values less than 2. An approximation is obtained by truncating the series. While it is invalid for general matrices, this approximation has been shown useful when used in the context of gradient based methods \citep{lorraine2020optimizing}. To do so one preconditions the matrix $H$ with a suitably chosen large divisor.

Note that the first order approximation is linear in $H$ and along with automatic differentiation methods allows easy and efficient multiplication with any vector by the Hessian-vector product (HVP) method  \citep{christianson1992automatic}. To compute the Hessian-vector product (HVP) with the vector $v$, one simply changes the parameters by $\epsilon v$ (for some small $\epsilon$) and computes the gradient. The difference between the two gradient when scaled equals the HVP. Furthermore this also holds when multiplying with the cross-Hessian $\partial_\theta \partial_\phi$, the same trick can be used once again. Next the terms in the von-Neumann series can be iteratively obtained by using HVP with the output of the previous iteration. This allows us to compute the series approximation to as many orders as desired. For further details refer to \citet{christianson1992automatic}

\section{Dataset Details}

\begin{table}
\centering
\begin{minipage}[b]{0.48\textwidth}\centering
\begin{tabular}{|l|lll|l|}
\hline
Dataset   & Train & Valid & Test & Label \\
\hline
Bibtex    & 4407  & 1491  & 1497 & 159   \\
Delicious & 9690  & 3207  & 3194 & 983   \\
Eurlexev  & 11557 & 3876  & 3881 & 3993 \\
\hline
\end{tabular}
\caption{Summary statistics for multi-label classification datasets }
\end{minipage}
\hfill
\begin{minipage}[b]{0.48\textwidth}\centering
\begin{tabular}{|l|l|l|ll|}
\hline
Dataset   & Size & Label & Avg Length & Avg Labels \\
\hline
AAPD    & 55840  & 54  & 163.4& 2.4 \\
RCV & 804414   & 103  &123.9 & 3.2\\
\hline
\end{tabular}
\caption{Summary statistics for large classification datasets}
\end{minipage}
\caption{Summary statistics \label{tab:apx_sum}}
\end{table}

For the larget textual datsets, we follow the processing of \citep{yang2018sgm} to preprocess the datasets. We filtered the dataset to 50000 words, and any texts longer than 500 words were discarded. For the smaller MLC datasets we used the standard splits. The details for both are presented in Table \ref{tab:apx_sum}.

For (POS) tagging, we follow \citep{tu2020improving} and use annotated datset from \citep{owoputi-etal-2013-improved}. The data set has 25 output tags. We also conduct experiments with small scale image segmentation on the Weizmann horses dataset \citep{borenstein2002class}. This is a classic dataset for structured prediction evaluation. It contains 328 images of horses and their manually labelled segmentation masks. For this task we follow the protocol detailed in \citet{lu2020structured}. 

\section{Analysis of Learnt Energies}

\subsection{MLP Classification}
\begin{figure}[htb]
    \centering
    \begin{subfigure}[b]{0.48\textwidth}
    \includegraphics[width=0.95\textwidth]{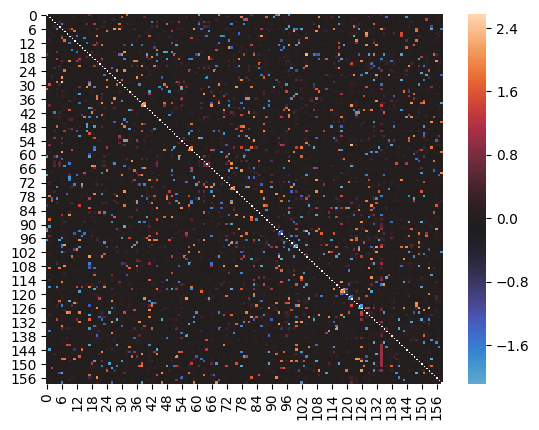}
    \caption{Energy Hessian}
    \label{fig:apx:bib}
    \end{subfigure}
    \begin{subfigure}[b]{0.46\textwidth}
     \centering
     \includegraphics[width=0.95\textwidth]{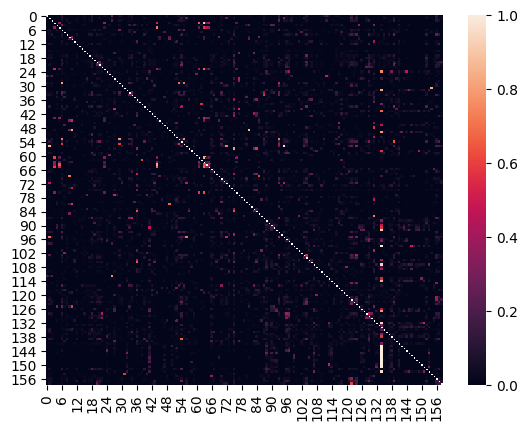}
     \caption{Co-occurence Matrix \label{fig:apx:coc}}
     \end{subfigure}
     \caption{Learnt energy and Label Cooccurence on bibtex \label{fig:apx:bib_heat_map}}
\end{figure}

The role of the global energy function $v^T\sigma(My)$ in $E_\phi$ is to model interaction between labels. The gradient of $E$ wrt the label $y$ has contributions from values of other labels, and optimization of $E$ should correspondingly increase or decrease the likelihood of a label, given the current probabilities of other labels. To test this hypothesis we compare the Hessian of the learned global energy wrt the output $y$. For frequently co-occurring labels $y_i,y_j$, increasing $y_i$ should give positively impact the gradient of $E$ wrt $y_j$. Similarly for pairs which co-occur less frequently the hessian should give negative values. In Figures \ref{fig:apx:bib_heat_map} we plot the average Hessian of the energy function over the instances as well as the co-occurence matrix of the labels. Note that the diagonals have been removed. We see a general correspondence between the co-occurence matrix and the hessian, though there are values which do not correspond. 

\subsection{POS Tagging}

\begin{figure}
    \centering
    \includegraphics[width=0.7\textwidth]{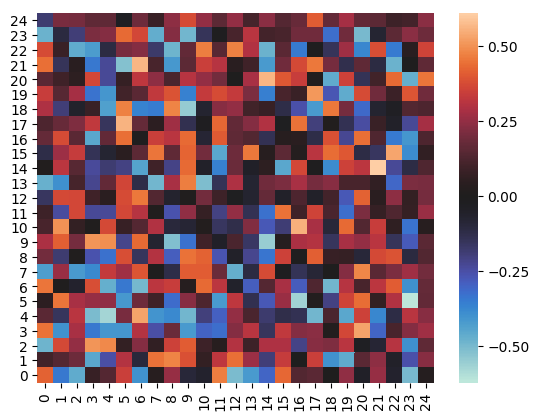}
    \caption{Heatmap representation of learnt pair correlation energies for POS tagging}
    \label{fig:apx:postag_heat_map}
\end{figure}

In Figure \ref{fig:apx:postag_heat_map} we plot average cross-Hessian of the energy function $E$ wrt $y_t,y_{t+1}$ in the POS tagging experiment. That is we are plotting $H = \dfrac{\partial^2 E}{\partial y_t \partial y_{t+1}}$.  Ideally the model should learn to put weight on tag pairs which follow each other i.e. if the tag A frequently appears after tag B, the term in $H_{\text{BA}}$ should be higher.  
Our experiments suggests this to be the case. As an example the tag 0 corresponds to nouns, 5 to verbs and 6 corresponds to adjectives. We can see from the map that the matrix downweights a noun-adjective and adjective-verb pairing; while upweighing the adjective-noun and noun-verb pairs.

\section{Additional Details}
\label{apx:more_expts}

\paragraph{Why Implicit Gradient and connection to Dynamic Losses}
The crucial difference between our method with $\mathcal{L}_{SSVM}$ loss and the InfNet+ proposal of \citet{tu2020improving} is the use of implicit gradient method  rather than alternative independent optimization. Note that training via independent optimization updates $\phi$ with only $\partial_\phi \LP$. From Equation \ref{eqn:grad_thm}, we can see that $\partial_\phi \LP$ is only one component of the true gradient, which includes an additional term (Line 1 of Equation \ref{eqn:grad_thm}). This additional term captures the indirect effect on $\LP$ due to $\theta$ being dependent on $\phi$ as well.  
If one uses the correct gradient, the optimization for energy ($\phi$) becomes explicitly aware and receives feedback from the inner optimization of the inference network. 
To highlight a difference form the standard EBM training procedure of \citet{tu2020improving}, consider a situation where the energy $E$ parameters have fully optimized $\LP$ ( for example margin loss). However, at the current parameters $\theta$ of the inference net $A$, the energy of the output $E(x, A(x))$ has a singular hessian $\partial^2_\theta E_\phi(x, A_\theta(x))$. 
The condition number of the Hessian is generally a good proxy for hardness of a convex optimization (e.g. gradient descent-based optimization will be slow due to the flatness of surface). Hence it is reasonable to suggest that finding the optimal $y$ is hard for the inference net . Under standard EBM training, the energy network will ignore this (as the updates depend only on $\partial_\phi \LP$) and stop updating. However, we can observe the first term in Equation \ref{eqn:grad_thm} that the first term depends on inverse hessian of the loss wrt $\theta$. As such we would see a a large gradient allowing the energy function to update itself.
Note furthermore that, in this sense, it is a learned dynamic loss \citep{wu2018learning, Bechtle19} which is actively adapting itself to the prediction networks behaviour.


\paragraph{Addition of task-loss in contrastive divergence}
The standard constrastive divergence \citep{hyvarinen2005estimation} depends only on the energy of positive and negative samples. In this case however we have added the task-loss $s$ to the negative samples, as it more strongly penalizes the energy of model outputs with high task loss. If we consider only one negative sample, then $\mathcal{L}_{CD}$ reduces to:
$$
 \log \dfrac{1}{1 + \exp(E_\phi(x,y) - E_\phi(x,\bar{y}_1) +  s(\bar{y}_k, y))  )} = \text{softplus}( s(\bar{y}_k, y) + E_\phi(x,y) - E_\phi(x,\bar{y}_1))
$$

which is essentially the same as the margin loss $\mathcal{L}_{SSVM}$ with the hinge being replaced by softer function. Furthermore similar to NCE\citep{gutmann2010noise} and CD\citep{hyvarinen2005estimation} losses, by using multiple samples it can provide greater information for the energy function, which is not possible for the SSVM loss.


\section{Additional Experiments}

Following \citet{tu2019benchmarking} we use our proposed method for a POS-tagging task as well. We use the Twitter-POS data from \citet{owoputi-etal-2013-improved}. The energy model $E_\phi$ is similar to the one used for NER tasks. We compare against a Bi-LSTM and CRF baseline, SPEN \citep{End-to-EndSPEN} and InfNet \citep{tu2019benchmarking} based model. Our results are presented in Table \ref{tab:struct_expt_pos1}. 

\begin{table}[hbt]
\centering
\begin{tabular}{|l||l|}
\hline
Models  & Accuracy  \\
\hline\hline
BILSTM  & 88.7  \\
SPEN   & 88.6 \\
CRF    & 89.3  \\
InfNet  & 89.7  \\
\hline
$\mathcal{L}_\text{CD}$     & \textbf{90.1} $\dagger$  \\
\hline
\end{tabular}
\caption{Test results for POS Tagging with different energy based models. SSVM and CD refers to our models. $\dagger$ indicates statistically significant \label{tab:struct_expt_pos1}}
\end{table}

\begin{table}[hbt]
\centering
\begin{tabular}{l|l}
\hline
Model & Mean IoU (\%)\\
\hline
\hline
DVN   & 83.9    \\
cGLOW & 81.2    \\
ALEN  & 85.7    \\
\hline
$\mathcal{L}_\text{CD}$   & \textbf{89.4} $\dagger$  \\
\hline
\end{tabular}
\caption{Mean IoU results for segmentation with different models on the Weizmann Horses dataset. The input image size is 32x32 pixels. $\dagger$ indicates statistically significant \label{tab:weisman_horse}}
\end{table}

We also conduct experiments with binary image segmentation on the Weizmann horses dataset. Following \citet{pan2020adversarial} and \citet{} we resize the images and masks to be 32 × 32 pixels. For the inference network we used a convolutional model with the same design as in cGLOW \citep{lu2020structured} like . As baselines we compare with DVN \citep{gygli2017deep}, ALEN \citep{pan2020adversarial}, and cGLOW \citep{lu2020structured}. The task loss in this case the the IoU (intersection over union) metric. Our results are reported in Table \ref{tab:weisman_horse}.
It is clear that our proposed method achieves the highest
IoU among the comparison methods with a close to 4\% percent improvement over ALEN, which itself is much ahead of other models. 
\end{document}